\documentclass{article}

\usepackage[nonatbib,preprint]{neurips_2020}

\usepackage{amssymb, amsmath, amsthm}

\usepackage{nccmath}

\usepackage{microtype}
\usepackage[english]{babel}
\usepackage{enumitem}
\usepackage{wrapfig}
\usepackage{multicol}
\usepackage{float}
\usepackage{adjustbox}
\usepackage{pdfrender}
\usepackage[theorems,skins]{tcolorbox}

\usepackage[utf8]{inputenc} 
\usepackage[T1]{fontenc}
\usepackage{amsfonts}
\usepackage{pifont}
\usepackage{bm}
\usepackage{bbm}
\usepackage{times}

\usepackage{fancyhdr}

\usepackage{graphicx}
\usepackage[labelfont=bf]{caption}
\usepackage[format=hang]{subcaption}

\usepackage{booktabs}

\usepackage[algoruled, algo2e]{algorithm2e}
\usepackage{algorithm}
\usepackage{algorithmic}
\usepackage{listings}
\usepackage{fancyvrb}
\fvset{fontsize=\normalsize}

\usepackage[numbers]{natbib}

\usepackage[acronym,smallcaps,nowarn]{glossaries}
\glsdisablehyper

\usepackage[colorlinks=true,linktoc=all]{hyperref}
\usepackage[all]{hypcap}
\hypersetup{citecolor=deepblue}
\hypersetup{linkcolor=deepblue}

\usepackage{url}

\usepackage[nameinlink]{cleveref}

\usepackage{todonotes}

\usepackage{multirow}

\usepackage{nicefrac}
\newacronym{crt}{crt}{conditional randomization test}
\newacronym{snp}{snp}{single nucleotide polymorphism}
\newacronym{fdr}{fdr}{false discovery rate}
\newacronym{fdp}{fdp}{false discovery proportion}
\newacronym{mle}{mle}{maximum likelihood estimation}
\newacronym{kld}{kl}{Kullback–Leibler divergence}
\newacronym{pt}{pt}{permutation test}
\newacronym{shap}{shap}{shapley additive explanations}
\newacronym{roc}{roc}{receiver operating characteristic}
\newacronym{rbf}{rbf}{radial basis function}
\newacronym{rf}{rf}{random forest}
\newacronym{ddlk}{ddlk}{deep direct likelihood knockoffs}
\newacronym{hrt}{hrt}{holdout randomization test}
\newacronym{ourmodel}{ddlk}{deep direct likelihood knockoffs}
\newacronym{gan}{gan}{generative adversarial network}
\newacronym{mmd}{mmd}{maximum mean discrepancy}
\newacronym{cbc}{cbc}{complete blood count}
\newacronym{vae}{vae}{variational auto-encoder}
\newacronym{hmm}{hmm}{hidden markov model}

\newtheorem{theorem}{Theorem}[section]

\newtheorem{lemma}[theorem]{Lemma}

\crefname{thmdef}{def.}{defs.}
\Crefname{thmdef}{Definition}{Definitions}
\crefname{lemma}{lemma}{lemmas}
\crefname{prop}{prop.}{props.}
\crefname{algorithm}{algorithm}{algorithms}
\Crefname{algorithm}{Algorithm}{Algorithms}
\crefname{algocf}{Alg.}{Algs.}
\Crefname{algocf}{Algorithm}{Algorithms}
\crefname{appsec}{app.}{apps.}
\Crefname{appsec}{A.}{A.}

\newif\ifprintcomments

\definecolor{deepblue}{HTML}{20639B}
\definecolor{seagreen}{HTML}{3CAEA3}
\definecolor{niceyellow}{HTML}{F6D55C}

\def\eqref#1{equation~\ref{#1}}

\def\1{\bm{1}}

\def\rvb{{\mathbf{b}}}

\def\rvu{{\mathbf{u}}}
\def\rvut{{\widetilde{\mathbf{u}}}}

\def\rvw{{\mathbf{w}}}
\def\rvx{{\mathbf{x}}}
\def\rvxt{{\widetilde{\mathbf{x}}}}
\def\rvy{{\mathbf{y}}}
\def\rvz{{\mathbf{z}}}

\def\vu{{\bm{u}}}
\def\vv{{\bm{v}}}

\def\vx{{\bm{x}}}
\def\vy{{\bm{y}}}

\DeclareMathAlphabet{\mathsfit}{\encodingdefault}{\sfdefault}{m}{sl}
\SetMathAlphabet{\mathsfit}{bold}{\encodingdefault}{\sfdefault}{bx}{n}

\newcommand{\E}{\mathbb{E}}

\newcommand{\KL}{{\mathrm{KL}}}

\DeclareMathOperator*{\argmax}{arg\,max}

\newcommand{\kl}[2]{\text{KL}(#1 \parallel #2)}
\newcommand{\vxt}{\widetilde{\vx}}
\newcommand{\vut}{\widetilde{\vu}}

\newcommand{\D}{\mathcal{D}}

\newcommand{\Djnt}{\widetilde{\D}_{j,N}}
\newcommand{\Dnt}{\widetilde{\D}_{N}}

\newcommand{\swap}[1]{\text{swap}(#1)}

\newcommand{\qjoint}{\hat{q}_{\text{joint}}}
\newcommand{\qknockoff}{\hat{q}_{\text{knockoff}}}
\newcommand{\qgumbel}{\hat{q}_{\text{gumbel}}}
\newcommand{\qresponse}{\hat{q}_{\text{response}}}

\newcommand{\cmark}{\ding{51}}%
\newcommand{\xmark}{\ding{55}}%

\title{\Acrlong{ddlk}}

\author{%
  Mukund Sudarshan\thanks{Corresponding author} \\
  Courant Institute of Mathematical Sciences\\
  New York University\\
  \texttt{sudarshan@cims.nyu.edu} \\
   \And
   Wesley Tansey \\
   Department of Epidemiology and Biostatistics \\
   Memorial Sloan Kettering Cancer Center \\
   \texttt{tanseyw@mskcc.org} \\
   \And
   Rajesh Ranganath \\
   Courant Institute of Mathematical Sciences\\
   Center for Data Science \\
   New York University\\
   \texttt{rajeshr@cims.nyu.edu} \\
}

\begin{document}

\maketitle

\begin{abstract}
Predictive modeling often uses black box machine learning methods, such as deep neural networks, to achieve state-of-the-art performance.
In scientific domains, the scientist often wishes to discover which features are actually important for making the predictions.
These discoveries may lead to costly follow-up experiments and as such it is important that the error rate on discoveries is not too high.
Model-X knockoffs \cite{ModelX_Knockoffs} enable important features to be discovered with control of the \gls{fdr}.
However, knockoffs require rich generative models capable of accurately modeling the knockoff features while ensuring they obey the so-called ``swap'' property.
We develop Deep Direct Likelihood Knockoffs (\acrshort{ddlk}), which directly minimizes the $\KL$ divergence implied by the knockoff swap property.
\acrshort{ddlk} consists of two stages: it first maximizes the explicit likelihood of the features, then minimizes the $\KL$ divergence between the joint distribution of features and knockoffs and any swap between them.
To ensure that the generated knockoffs are valid under any possible swap, \acrshort{ddlk} uses the Gumbel-Softmax trick to optimize the knockoff generator under the worst-case swap.
We find \acrshort{ddlk} has higher power than baselines while
controlling the false discovery rate on a variety of synthetic and real benchmarks including a task involving a large dataset from one of the epicenters of COVID-19.

\end{abstract}

\section{Introduction}
High dimensional multivariate datasets pervade many disciplines including biology, neuroscience, and medicine.
In these disciplines, a core question is which variables are important to predict the phenomenon being observed.
Finite data and noisy observations make finding informative variables impossible without some error rate.
Scientists therefore seek to find informative variables subject to a specific tolerance on an error rate such as the false discovery rate (\gls{fdr}).

Traditional methods to control to the \gls{fdr} rely on assumptions on how
the covariates of interest $\rvx$ may be related to the response $\rvy$.
Model-X knockoffs \citep{ModelX_Knockoffs} provide an alternative framework that controls the \gls{fdr} by constructing
synthetic variables $\rvxt$, called  knockoffs, that look like the original
covariates, but have no relationship with the response given the original covariates.
Variables of this form can be used to test the conditional independence of
each covariate in a collection, with the response
given the rest of the covariates by comparing the association the original covariate has
with the response with the association the knockoff has with the response.

The focus on modeling the covariates
shifts the testing burden to building good generative models of the covariates.
Knockoffs need to satisfy two properties: (i) they need to be independent of the response given the real covariates, (ii) they need to be equal in distribution when any subset of variables is swapped between knockoffs and real data.
Satisfying the first property is trivial by generating the knockoffs without looking using the response.
The second property requires building conditional generative models that are able
to match the distribution of the covariates.

\paragraph{Related work.}

Existing knockoff generation methods can be broadly classified as either model-specific or flexible.
Model-specific methods such as \glspl{hmm} \citep{sesia2017gene} or Auto-Encoding Knockoffs \citep{liu2018auto} make assumptions about the covariate distribution, which can be problematic if the data does not satisfy these assumptions.
\Glspl{hmm} assume the joint distribution of the covariates can be factorized into a markov chain.
Auto-Encoding Knockoffs use \glspl{vae} to model $\rvx$ and sample knockoffs $\rvxt$.
\Glspl{vae} assume $\rvx$ lies near a low dimensional manifold, whose dimension is controlled by a latent variable.
Covariates that violate this low-dimensional assumption can be better modeled by increasing the dimension of the latent variable, but risk retaining more information about $\rvx$, which can
reduce the power to select important variables.

Flexible methods for generating knockoffs such as KnockoffGAN \citep{knockoffgan} or Deep Knockoffs \citep{DeepKnockoffs} focus on likelihood-free generative models.
KnockoffGAN uses \gls{gan}-based generative models, which can be difficult to estimate \citep{mescheder2018training} and
sensitive to hyperparameters \citep{salimans2016improved, wgan-train,mescheder2017numerics}.
Deep Knockoffs employ \glspl{mmd}, the effectiveness of which often depends on the choice of a kernel which can involve selecting a bandwidth hyperparameter.
\citet{ramdas2015decreasing} show that in several cases, across many choices of bandwidth, \gls{mmd} approaches 0 as dimensionality increases while $\KL$ divergence remains non-zero, suggesting \glspl{mmd} may not reliably generate high-dimensional knockoffs.
Deep Knockoffs also prevent the knockoff generator from memorizing the covariates by explicitly controlling the correlation between the knockoffs and covariates.
This is specific to second order moments, and may ignore higher order ones present in the data.

We propose \gls{ddlk}, a likelihood-based method for generating knockoffs without the use of latent variables. 
\Gls{ddlk} is a two stage algorithm.
The first stage uses maximum likelihood
to estimate the distribution of the covariates from observed data.
The second stage estimates the knockoff distribution with likelihoods by minimizing the \gls{kld} between the joint distribution of the real covariates with knockoffs and the joint distribution of any swap of coordinates between covariates and knockoffs.
\Gls{ddlk} expresses the likelihoods for swaps in terms of the original
joint distribution with real covariates swapped with knockoffs.
Through the Gumbel-Softmax trick \citep{gumbel-jang, gumbel-maddison}, we optimize the knockoff distribution under the worst swaps.
By ensuring that the knockoffs are valid in the worst cases, \gls{ddlk} learns valid knockoffs in all cases.
To prevent \gls{ddlk} from memorizing covariates, we introduce a regularizer to encourage high conditional entropy for the knockoffs given the covariates.
We study \gls{ddlk} on synthetic, semi-synthetic, and real datasets.
Across each study, \gls{ddlk} controls the \gls{fdr} while achieving higher power than competing \gls{gan}-based, \gls{mmd}-based, and autoencoder-based methods.

\vspace{-0.2cm}
\section{Knockoff filters}
\vspace{-0.2cm}
\label{sec:kfilters}

Model-X knockoffs \citep{ModelX_Knockoffs} is a tool used to build variable selection methods.
Specifically, it facilitates the control of the \gls{fdr}, which is the proportion of selected variables that are not important.
In this section, we review the requirements to build variable selection methods using Model-X knockoffs.

Consider a data generating distribution $q(\rvx) q(\rvy \mid \rvx)$ where variables $\rvx \in \mathbb{R}^d$, response $\rvy$ only depends on $\rvx_S$,  $S \subseteq [d]$, and $\rvx_S$ is a subset of the variables.
Let $\hat{S}$ be a set of indices identified by a variable selection algorithm.
The goal of such algorithms is to return an $\hat{S}$ that maximizes the number of indices in $S$ while maintaining the \gls{fdr} at some nominal level:
\begin{align*}
    \text{\gls{fdr}} = \E \left[\frac{ \vert \{j: j \in \hat{S} \setminus S\} \vert }{ \max \left( \vert \{j: j \in \hat{S}\} \vert , 1 \right) }\right] .
\end{align*}
To control the \gls{fdr} at the nominal level, Model-X knockoffs requires (a) knockoffs $\rvxt$, and (b) a knockoff statistic $w_j$ to assess the importance of each feature $\rvx_j$.

Knockoffs $\rvxt$ are random vectors that satisfy the following properties for any set of indices $H \subseteq [d]$:
\begin{align}
[\rvx, \rvxt] & \overset{d}{=} [\rvx, \rvxt]_{\text{swap}(H)} \label{eqn:swap} \\
\rvy & \perp \rvxt \mid \rvx. \label{eqn:conditional-independence}
\end{align}
The swap property \cref{eqn:swap} ensures that the joint distribution of $[\rvx, \rvxt]$ is invariant under any swap.
A swap operation at position $j$ is defined as exchanging the entry of $\rvx_j$ and $\rvxt_j$.
For example, when $\rvx = [\rvx_1, \rvx_2, \rvx_3]$, and $H = \{1, 3\}$, $[\rvx, \rvxt]_{\text{swap}(H)} = [\rvxt_1, \rvx_2, \rvxt_3, \rvx_1, \rvxt_2, \rvx_3]$.
\Cref{eqn:conditional-independence} ensures that the response $\rvy$ is independent of the knockoff $\rvxt$ given the original features $\rvx$.

A knockoff statistic $w_j$ must satisfy the flip-sign property.
This means if $\rvx_j \in S$, $w_j$ must be positive.
Otherwise, the sign of $w_j$ must be positive or negative with equal probability.

Given knockoff $\rvxt$ and knockoff statistics $\{ w_j \}_{j=1}^d$, exact control of the \gls{fdr} at level $p$ can be obtained by selecting variables in $\rvx$ such that $w_j > \tau_p$.
The threshold $\tau_p$ is given by:
\begin{align}
    \tau_p = \underset{t > 0}{\min}\left\{t : \frac{1 + | \{j: w_j \leq -t \} | }{ | \{j: w_j \geq t \} |  } \leq p \right \}\label{eq_fdr_control}.
\end{align}

While knockoffs are a powerful tool to ensure that the \gls{fdr} is controlled at the nominal level, the choice of method to generate knockoffs is left to the practitioner.

Existing methods for knockoffs include model-specific approaches
that make specific assumptions about the covariate distribution, and flexible likelihood-free methods.
If the joint distribution of $\rvx$ cannot be factorized into a markov chain \citep{sesia2017gene} or if $\rvx$ does not lie near a low-dimensional manifold \citep{liu2018auto}, model-specific generators will yield knockoffs that are not guaranteed to control the \gls{fdr}.
Likelihood-free generation methods that use \glspl{gan} \citep{knockoffgan} or \glspl{mmd} \citep{DeepKnockoffs} make fewer assumptions about $\rvxt$, but can be difficult to estimate \citep{mescheder2018training}, sensitive to hyperparameters \citep{salimans2016improved,wgan-train,mescheder2017numerics}, or suffer from low power in high dimensions \citep{ramdas2015decreasing}.
In realistic datasets, where $\rvx$ can come from an arbitrary distribution and dimensionality is high, it remains to be seen how to reliably generate knockoffs that satisfy \cref{eqn:swap,eqn:conditional-independence}.

\vspace{-0.2cm}
\section{\Acrlong{ddlk}}
\vspace{-0.2cm}

We motivate \gls{ddlk} with the following observation.
The swap property in \cref{eqn:swap} is satisfied if the $\KL$ divergence between the
original and swapped distributions is zero.
Formally,
let $H$ be a set of indices to swap,
and $\rvz = [\rvx, \rvxt]$, $\rvw = [\rvx, \rvxt]_{\swap{H}}$.
Then under any such $H \subseteq [d]$:
\begin{align}
	\kl{q_\rvz}{q_\rvw} 
	= 
	\E_{q_\rvz(\rvz)}
	 \left[ \log \frac{q_\rvz(\rvz)}{q_\rvw(\rvz)} \right] = 0. \label{eqn:ddlk-motivation}
\end{align}

A natural algorithm for generating valid knockoffs might be to parameterize each distribution above and solve for the parameters by minimizing the LHS of \cref{eqn:ddlk-motivation}.
However, modeling $q_\rvw$ for every possible swap is difficult and computationally infeasible in high dimensions.
\Cref{thm:swap} provides a useful solution to this problem.

\begin{theorem}
\label{thm:swap}
Let $\mu$ be a probability measure defined on a measurable space.
Let $f_H$ be a swap function using indices $H \subseteq [d]$.
If $\vv$ is a sample from $\mu$, the probability law of $f_H(\vv)$ is $\mu \circ f_H$.
\end{theorem}

As an example, in the continuous case, where 
$q_\rvz$ and $q_\rvw$ are the densities of $\rvz$ and $\rvw$ respectively, $q_\rvw$ evaluated at a sample $\vv$ is simply $q_\rvz$ evaluated at the swap of $\vv$.
We show the direct proof of this example and \cref{thm:swap} in \cref{sec:proofs}.
A useful consequence of \cref{thm:swap} is that \gls{ddlk} needs to only model $q_\rvz$, instead of $q_\rvz$ and every possible swap distribution $q_\rvw$.
To derive the \gls{ddlk} algorithm, we first expand \cref{eqn:ddlk-motivation}:
\begin{align}
	\E_{q_\rvz(\rvz)}
	\left[ \log \frac{q_\rvz(\rvz)}{q_\rvw(\rvz)} \right]
	=
	\E_{q(\rvx)}
	\E_{q(\rvxt \mid \rvx)}
	\left[ \log \frac{q(\rvx) q(\rvxt \mid \rvx)}{q(\rvu) q(\rvut \mid \rvu)} \right] \label{eqn:ddlk-kl},
\end{align}
where $[\rvu, \rvut] = [\rvx, \rvxt]_{\swap{H}}$.
\Gls{ddlk} models the RHS by parameterizing $q(\rvx)$ and $q(\rvxt \mid \rvx)$ with $\qjoint(\rvx ; \theta)$ and $\qknockoff(\rvxt \mid \rvx ; \phi)$ respectively.
The parameters $\theta$ and $\phi$ can be optimized separately in two stages.

\paragraph{Stage 1: Covariate distribution estimation.}

We model the distribution of $\rvx$ using $\qjoint(\rvx ; \theta)$.
The parameters of the model $\theta$ are learned by maximizing $\E_{\vx \sim \D_N} \left[ \log \qjoint(\vx; \theta) \right]$ over a dataset $\D_N := \{ \vx^{(i)} \}_{i=1}^N$ of $N$ samples.

\paragraph{Stage 2: Knockoff generation.}

For any fixed swap $H$, minimizing the KL divergence between the following distributions ensures the swap property \cref{eqn:swap} required of knockoffs:
\begin{align}
	 \kl{\qjoint(\rvx ; \theta) \qknockoff(\rvxt \mid \rvx ; \phi)}{\qjoint(\rvu ; \theta) \qknockoff(\rvut \mid \rvu ; \phi)}.
	 \label{eqn:swap-loss}
\end{align}
Fitting the knockoff generator $\qknockoff(\rvxt \mid \rvx ; \phi)$ involves minimizing this KL divergence for all possible swaps $H$.
To make this problem tractable, we use several building blocks that help us (a) sample swaps with the highest values of this KL and (b) prevent $\qknockoff$ from memorizing $\rvx$ to trivially satisfy the swap property in \cref{eqn:swap}.

\subsection{Fitting \gls{ddlk}}

Knockoffs must satisfy the swap property \cref{eqn:swap} for all potential sets of swap indices $H \subseteq [d]$.
While this seems to imply that the KL objective in \cref{eqn:swap-loss} must be minimized under an exponential number of swaps, swapping every coordinate suffices \citep{candes2018panning}.
More generally, showing the swap property for a
collection of sets where every coordinate 
can be represented as the symmetric difference of members of the collection
is sufficient.
See \cref{sec:sufficient-swap} more more details.

\paragraph{Sampling swaps.}

Swapping $d$ coordinates can be expensive in high dimensions, so existing methods resort to randomly sampling swaps \citep{DeepKnockoffs,knockoffgan} during optimization.
Rather than sample each coordinate uniformly at random, we propose parameterizing the sampling process for swap indices $H$ so that swaps sampled from this process 
yields large values of the KL objective in \cref{eqn:swap-loss}.
We do so because of the following property of swaps, which we prove in \cref{sec:proofs}.
\begin{lemma}
\label{lemma:gumbel-correctness}
\textbf{Worst case swaps}:
Let $q(H ; \beta)$ be the worst case swap distribution.
That is, the distribution over swap indices that maximizes
\begin{align}
	\E_{H \sim q(H ; \beta)} \kl{\qjoint(\rvx ; \theta) \qknockoff(\rvxt \mid \rvx ; \phi)}{\qjoint(\rvu ; \theta) \qknockoff(\rvut \mid \rvu ; \phi)} \label{eqn:swap-loss-expected}
\end{align}
with respect to $\beta$.
If \cref{eqn:swap-loss-expected} is minimized with respect to $\phi$, knockoffs sampled from $\qknockoff$ will satisfy the swap property in \cref{eqn:swap} for any swap $H$ in the power set of $[d]$.
\end{lemma}
Randomly sampling swaps can be thought of as sampling from $d$ Bernoulli random variables $\{ \rvb_j \}_{j=1}^d$ with parameters $\beta = \{ \beta_j \}_{j=1}^d$ respectively, where each $\rvb_j$ indicates whether the $j$th coordinate is to be swapped.
A set of indices $H$ can be generated by letting $H = \{ j : \rvb_j = 1 \}$.
To learn a sampling process that helps maximize \cref{eqn:swap-loss-expected}, we optimize the values of $\beta$.
However, since score function gradients for the parameters of Bernoulli random variables can have high variance, \gls{ddlk} uses a continuous relaxation instead.
For each coordinate $j \in d$, \gls{ddlk} learns the parameters for a Gumbel-Softmax \citep{gumbel-jang,gumbel-maddison} distribution $\qgumbel(\beta_j)$.

\begin{algorithm}[t]
\caption{\gls{ddlk}}
\label{alg:ddlk}
\DontPrintSemicolon
\SetAlgoLined
\SetKwRepeat{Do}{do}{while}
\KwIn{$\D_N := \{ \vx^{(i)} \}_{i=1}^N$, dataset of covariates; $\lambda$, regularization parameter; $\alpha_\phi$, learning rate for $\qknockoff$; $\alpha_\beta$, learning rate for $\qgumbel$}
\KwOut{$\theta$, parameter for $\qjoint$, $\qknockoff$, parameter for $\qknockoff$}
$\theta = \argmax_{\theta} \frac{1}{N} \sum_{i=1}^N \log \qjoint(\vx^{(i)}; \theta)$\;

\While{$\qknockoff$ not converged}{
Sample $\{ \vxt^{(i)} \}_{i=1}^N$, where $\vxt^{(i)} \sim \qknockoff(\rvxt \mid \vx^{(i)}; \phi)$\;

Sample swap $H \sim \qgumbel(\beta)$\;

Create $\{ (\vu^{(i)}, \vut^{(i)}) \}_{i=1}^N$, where $[\vu^{(i)}, \vut^{(i)}] = [\vx^{(i)}, \vxt^{(i)}]_{\swap{H}}$\;

Let $\mathcal{A}(\phi) = \frac{1}{N} \sum_{i=1}^N \log \qjoint(\vx^{(i)}; \theta) + (1 + \lambda) \log \qknockoff(\vxt^{(i)} \mid \vx^{(i)}; \phi)$\;

Let $\mathcal{B}(\phi, \beta) = \frac{1}{N} \sum_{i=1}^N \log \qjoint(\vu^{(i)}; \theta) +\log \qknockoff(\vut^{(i)} \mid \vu^{(i)}; \phi)$\;

$\phi \gets \phi - \alpha_\phi \nabla_\phi (\mathcal{A}(\phi) - \mathcal{B}(\phi, \beta))$\;

$\beta \gets \beta + \alpha_\beta \nabla_\beta (\mathcal{A}(\phi) - \mathcal{B}(\phi, \beta))$\;
}

\Return{$\theta$, $\phi$, $\beta$}
\end{algorithm}

\paragraph{Entropy regularization.}

Minimizing the the KL objective in \cref{eqn:swap-loss} over the worst case swap distribution will generate knockoffs that satisfy the swap property \cref{eqn:swap}.
However,
a potential solution in the optimization of $\qknockoff(\rvxt \mid \rvx)$ is to memorize the covariates $\rvx$, which reduces the power to select important variables.

To solve this problem, \gls{ddlk} introduces a regularizer based on the conditional entropy, to push $\rvxt$ to not be a copy of $\rvx$.
This regularizer takes the form $-\lambda \E[- \log \qknockoff(\rvxt \mid \rvx ; \phi)]$, where $\lambda$ is a hyperparameter.

Including the regularizer on conditional entropy, and Gumbel-Softmax sampling of swap indices, the final optimization objective for \gls{ddlk} is:
\begin{align}
	\min\limits_{\phi} \max\limits_{\beta}
	\E_{H \sim \qgumbel(\beta)}
	\E_{\vx \sim \D_N}
	\E_{\vxt \sim \qknockoff(\rvxt \mid \vx ; \phi)}
	\log \frac{\qjoint(\vx ; \theta) \qknockoff(\vxt \mid \vx ; \phi)^{1 + \lambda}}{\qjoint(\vu ; \theta) \qknockoff(\vut \mid \vu ; \phi)}
	\label{eqn:ddlk-objective}
\end{align}
where $[\vu, \vut] = [\vx, \vxt]_{\swap{H}}$.
We show the full \gls{ddlk} algorithm in \cref{alg:ddlk}.
\Gls{ddlk} fits $\qjoint$ by maximizing the likelihood of the data.
It then fits $\qknockoff$ by optimizing \cref{eqn:ddlk-objective} with noisy gradients.
To do this, \gls{ddlk} first samples knockoffs conditioned on the covariates and a set of swap coordinates, then computes Monte-Carlo gradients of the \gls{ddlk} objective in \cref{eqn:ddlk-objective} with respect to parameters $\phi$ and $\beta$.
In practice \gls{ddlk} can use stochastic gradient estimates like the score function or reparameterization gradients for this step.
The $\qjoint$ and $\qknockoff$ models can be implemented with flexible models like MADE \citep{germain2015made} or mixture density networks \citep{bishop:1994:mixture-density-networks}.

\vspace{-0.2cm}
\section{Experiments}
\vspace{-0.2cm}
We study the performance of \gls{ddlk} on several synthetic, semi-synthetic, and real-world datasets.
We compare \gls{ddlk} with several non-Gaussian knockoff generation methods: 
Auto-Encoding Knockoffs (AEK) \citep{liu2018auto}, KnockoffGAN \citep{knockoffgan}, and Deep Knockoffs \citep{DeepKnockoffs}.\footnote{For all comparison methods, we downloaded the publicly available implementations of the code (if available) and used the appropriate configurations and hyperparameters recommended by the authors. 
See \cref{sec:baseline-model-settings} for code and model hyperparameters used.
}

\begin{figure}[t]
\includegraphics[width=\textwidth,page=1,trim=0 0 0 0,clip]{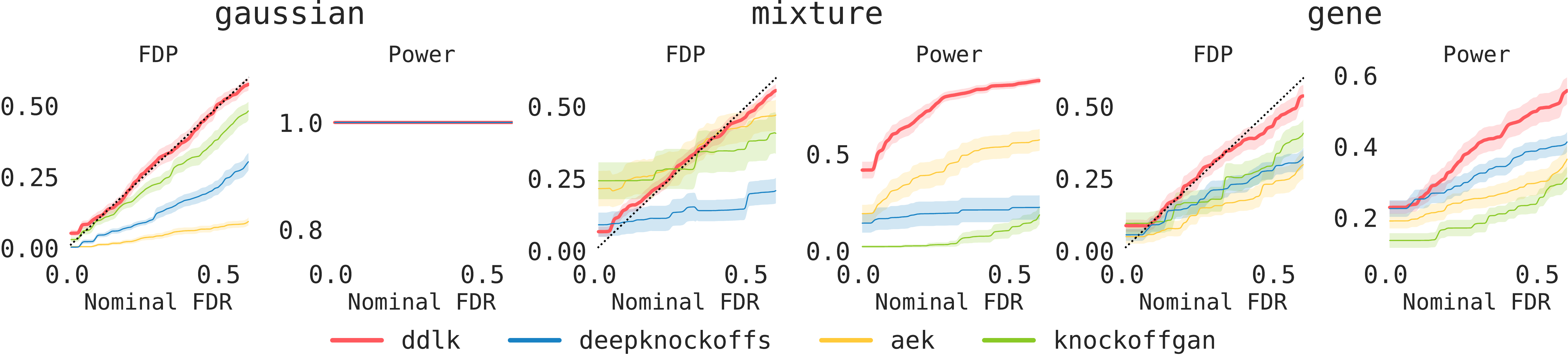}
\caption{
\textbf{\gls{ddlk} controls \gls{fdr} at the nominal rate and achieves highest power on a variety of benchmarks.}
For each benchmark, we show the \gls{fdp} and power of each knockoff method.
}
\label{fig:fdr-power-plots}
\vspace{-0.1cm}
\end{figure}

Each experiment involves three stages.
First, we fit a knockoff generator using a dataset of covariates $\{ \vx^{(i)} \}_{i=1}^N$.
Next, we fit a response model $\qresponse(\rvy \mid \rvx ; \gamma)$, and use its performance on a held-out set to create a knockoff statistic $w_j$ for each feature $\rvx_j$.
We detail the construction of these test statistics in \cref{sec:mixture-statistic}.
Finally, we apply a knockoff filter to the statistics $\{w_j\}_{j=1}^d$ to select features at a nominal \gls{fdr} level $p$, and measure the ability of each knockoff method to select relevant features while maintaining \gls{fdr} at $p$.
For the synthetic and semi-synthetic tasks, we repeat these three stages 30 times  to obtain interval estimates of each performance metric.

In our experiments, we assume $\rvx$ to be real-valued with continuous support and decompose the models $\qjoint$ and $\qknockoff$ via the chain rule:
$
q(\rvx \mid \cdot) = q(\rvx_1 \mid \cdot) \prod_{j = 2}^d q(\rvx_j \mid \cdot, \rvx_1, \dots, \rvx_{j-1}).
$
For each conditional distribution, we fit a mixture density network \cite{bishop:1994:mixture-density-networks} where the number of mixture components is a hyperparameter.
In principle, any model that produces an explicit likelihood value can be used to model each conditional distribution.

Fitting $\qjoint$ involves using samples from a dataset, but fitting $\qknockoff$ involves sampling from $\qknockoff$.
This is a potential issue because the gradient of the \gls{ddlk} objective \cref{eqn:ddlk-objective} with respect to $\phi$ is difficult to compute as it involves integrating $\qknockoff$, which depends on $\phi$.
To solve this, we use an implicit reparameterization \citep{figurnov2018implicit} of mixture densities.
Further details of this formulation are presented in \cref{sec:implicit}.

Across each benchmark involving \gls{ddlk}, we vary only the $\lambda$ entropy regularization parameter based on the amount of dependence among covariates.
The number of parameters, learning rate, and all other hyperparameters are kept constant.
To sample swaps $H$, we sample using a straight-through Gumbel-Softmax estimator \citep{gumbel-jang}.
This allows us to sample binary values for each swap, but use gradients of a continuous approximation during optimization.
For brevity, we present the exact hyperparameter details for \gls{ddlk} in \cref{sec:hyperparameters}.

We run each experiment on a single CPU with 4GB of memory.
\Gls{ddlk} takes roughly 40 minutes in total to fit both $\qjoint$ and $\qknockoff$ on a 100-dimensional dataset.

\paragraph{Synthetic benchmarks.}

Our tests on synthetic data seek to highlight differences in power and \gls{fdr} between each knockoff generation method.
Each dataset in this section consists of $N=2000$ samples, 100 features, 20 of which are used to generate the response $\rvy$.
We split the data into a training set ($70\%$) to fit each knockoff method, a validation set ($15\%$) used to tune the hyperparameters of each method, and a test set ($15\%$) for evaluating knockoff statistics.

[\texttt{gaussian}]: We first replicate the multivariate normal benchmark of  \citet{DeepKnockoffs}. We sample $\rvx \sim \mathcal{N}(0, \Sigma)$, where $\Sigma$ is a $d$-dimensional covariance matrix whose entries $\Sigma_{i,j} = \rho^{|i-j|}$.
This autoregressive Gaussian data exhibits strong correlations between adjacent features, and lower correlations between features that are further apart.
We generate $\rvy \mid \rvx \sim \mathcal{N}(\langle \rvx, \alpha \rangle, 1)$, where coefficients for the important features are drawn as $\alpha_j \sim  \frac{100}{\sqrt{N}} \cdot \text{Rademacher}(0.5)$.
In our experiments, we set $\rho=0.6$.
We let the \gls{ddlk} entropy regularization parameter $\lambda = 0.1$.
Our model $\qresponse$ for $\rvy \mid \rvx$ is a $1$-layer neural network with $200$ parameters.

\begin{figure*}[t]
    \centering
    \includegraphics[width=\textwidth,page=1,trim=0 0 0 0,clip]{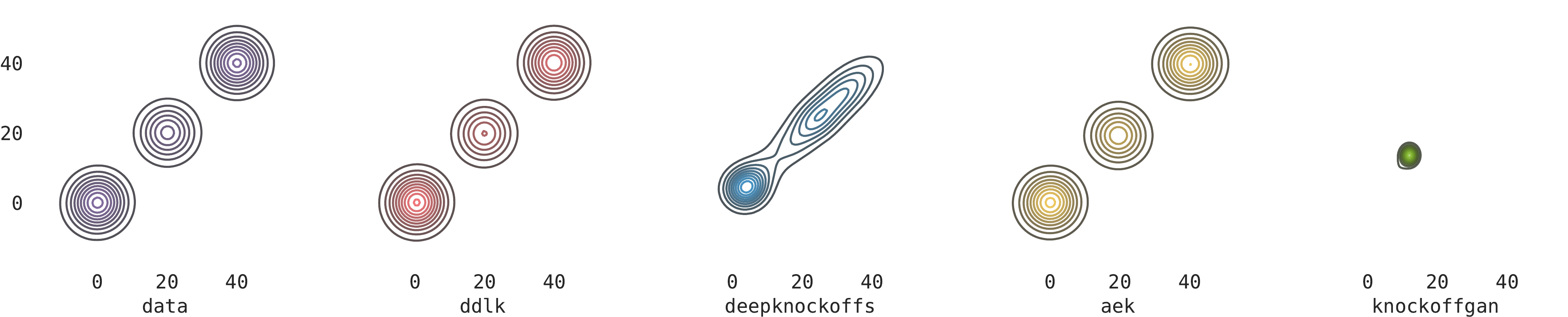}
    \caption{
   \textbf{\Gls{ddlk} closely models the modes, covariances, and mixture proportions of the \texttt{mixture} dataset}. Auto-Encoding Knockoffs also capture every mode, but does so by overfitting to the covariates. Deep Knockoffs are able to match the first two moments, but fail to capture every mode. KnockoffGAN suffers from mode collapse and fails to capture every mode.
    }
    \label{fig:gmm-joint}
    \vspace{-0.3cm}
\end{figure*}

[\texttt{mixture}]: To compare each method on its ability to generate non-Gaussian knockoffs, we use a mixture of autoregressive Gaussians.
This is a more challenging benchmark as each covariate is multi-modal, and highly correlated with others.
We sample $\rvx \sim \sum_{k=1}^K \pi_k \mathcal{N}(\mu_k, \Sigma_k)$, where each $\Sigma_k$ is a $d$-dimensional covariance matrix whose $(i,j)$th entry is $\rho_k^{|i-j|}$.
We generate $\rvy \mid \rvx \sim \mathcal{N}(\langle \rvx, \alpha \rangle, 1)$, where coefficients for the important features are drawn as  $\alpha_j \sim  \frac{100}{\sqrt{N}} \cdot \text{Rademacher}(0.5)$.
In our experiments, we set $K=3$, and $(\rho_1, \rho_2, \rho_3) = (0.6, 0.4, 0.2)$.
Cluster centers are set to $(\mu_1, \mu_2, \mu_3) = (0, 20, 40)$, and mixture proportions are set to $(\pi_1, \pi_2, \pi_3) = (0.4, 0.2, 0.4)$.
We let the \gls{ddlk} entropy regularization parameter $\lambda = 0.001$.
\Cref{fig:gmm-joint} visualizes two randomly selected dimensions of this data.
\vspace{-0.2cm}

\paragraph{\emph{Results.}}

\Cref{fig:fdr-power-plots} compares the average \gls{fdp} and power (percentage of important features selected) of each knockoff generating method. The average \gls{fdp} is an empirical estimate of the \gls{fdr}.
In the case of the \texttt{gaussian} dataset, all methods control \gls{fdr} at or below the the nominal level, while achieving 100\% power to select important features.
The main difference between each method is in the calibration of null statistics.
Recall that a knockoff filter assumes a null statistic to be positive or negative with equal probability, and features with negative statistics below a threshold
are used to control the number of false
discoveries when features with positive statistics above the same threshold are selected.
\Gls{ddlk} produces the most well calibrated null statistics as evidenced by the closeness of its \gls{fdp} curve to the dotted diagonal line.

\Cref{fig:fdr-power-plots} also demonstrates the effectiveness of \gls{ddlk} in modeling non-Gaussian covariates.
In the case of the \texttt{mixture} dataset, \gls{ddlk} achieves significantly higher power than the baseline methods, while controlling the \gls{fdr} at nominal levels.
To understand why this may be the case, we plot the joint distribution of two randomly selected features in \cref{fig:gmm-joint}.
\Gls{ddlk} and Auto-Encoding Knockoffs both seem to capture all three modes in the data.
However, Auto-Encoding Knockoffs tend to produce knockoffs that are very similar to the original features, and yield lower power when selecting variables, shown in \cref{fig:fdr-power-plots}.
Deep Knockoffs manage to capture the first two moments of the data -- likely due to an explicit second-order term in the objective function -- but tend to over-smooth and fail to properly estimate the knockoff distribution.
KnockoffGAN suffers from mode collapse, and fails to capture even the first two moments of the data.
This yields knockoffs that not only have low power, but also fail to control \gls{fdr} at nominal levels.

\paragraph{Robustness of \gls{ddlk} to entropy regularization.}

\begin{wrapfigure}{l}{0.5\textwidth}
\centering
  \includegraphics[width=0.9\linewidth,page=1,trim=0 0 0 0,clip]{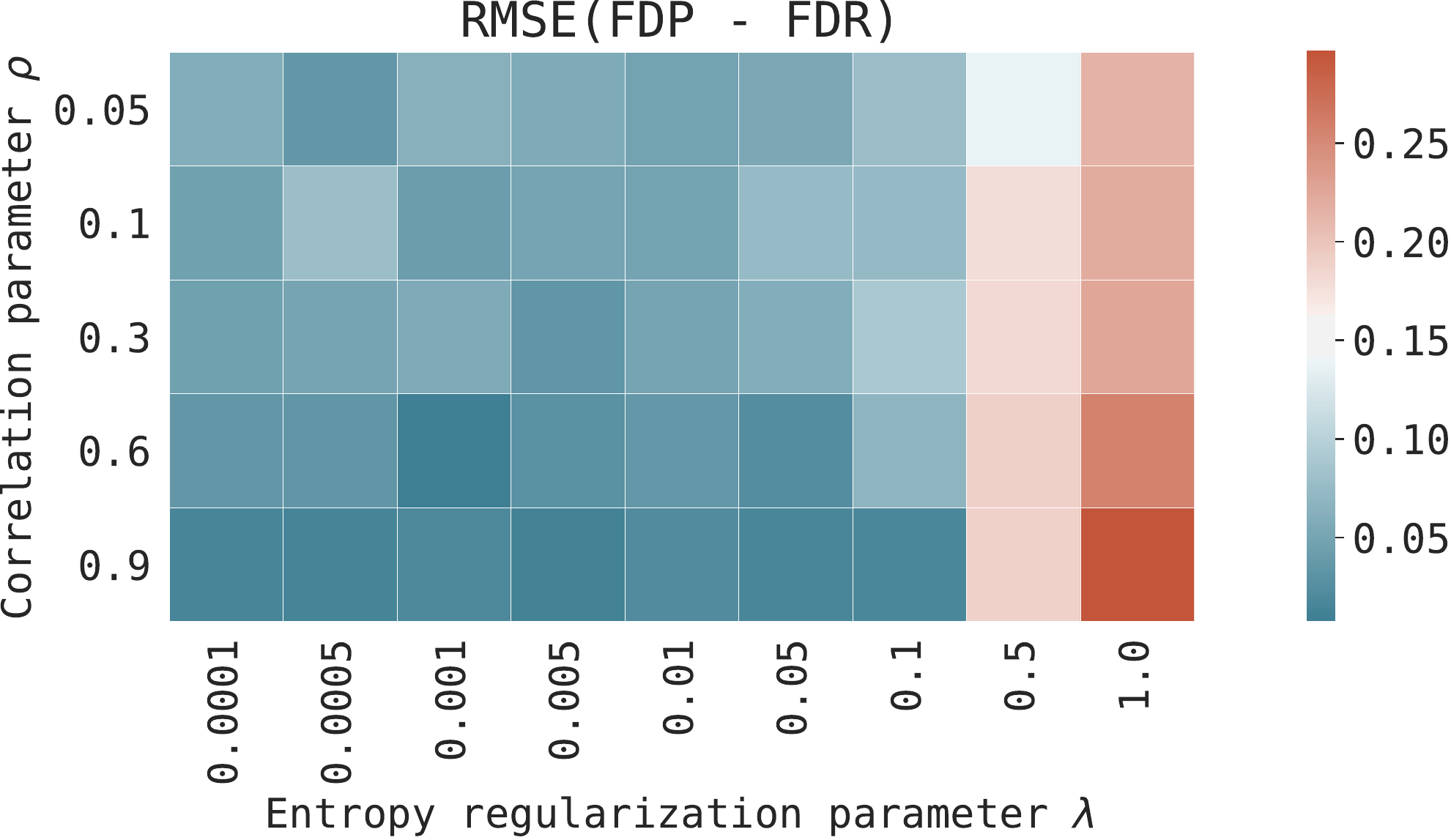}
  \caption{
  \textbf{\gls{ddlk} is robust to choices of entropy regularization parameter $\lambda$}.
  For most choices of $\lambda \leq 0.1$, \gls{ddlk} achieves \gls{fdp} very close to that of the nominal \gls{fdr} rate.
This figure shows the RMSE between the expected and actual \gls{fdp} curves.
  }
  \label{fig:reg_entropy}
  \vspace{-0.5cm}
\end{wrapfigure}

To provide guidance on how to set the entropy regularization parameter, we explore the effect of $\lambda$ on both \gls{fdr} control and power.
Intuitively, lower values of $\lambda$ will yield solutions of $\qknockoff$ that may satisfy \cref{eqn:swap} and control \gls{fdr} well, but may also memorize the covariates and yield low power.
Higher values of $\lambda$ may help improve power, but at the cost of \gls{fdr} control.
In this experiment, we again use the \texttt{gaussian} dataset, but vary $\lambda$ and the correlation parameter $\rho$.
\Cref{fig:reg_entropy} highlights the performance of \gls{ddlk} over various settings of $\lambda$ and $\rho$.
We show a heatmap where each cell represents the RMSE between the nominal \gls{fdr} and mean \gls{fdp} curves over 30 simulations.
In each of these settings \gls{ddlk} achieves a power of 1, so we only visualize \gls{fdp}.
We observe that the \gls{fdp} of \gls{ddlk} is very close to its expected value for most settings where $\lambda \leq 0.1$.
This is true over a wide range of $\rho$ explored, demonstrating that \gls{ddlk} is not very sensitive to the choice of this hyperparameter. We also notice that data with weaker correlations see a smaller increase in \gls{fdp} with larger values of $\lambda$.
In general, checking the \gls{fdp} on synthetic responses 
generated conditional on real covariates can aid in selecting $\lambda$.

\paragraph{Semi-synthetic benchmark.}

[\texttt{gene}]: In order to evaluate the \gls{fdr} and power of each knockoff method using covariates found in a genomics context, we create a semi-synthetic dataset.
We use RNA expression data of 963 cancer cell lines from the Genomics of Drug Sensitivity in Cancer study \citep{yang2012genomics}.
Each cell line has expression levels for 20K genes, of which we sample 100 such that every feature is highly correlated with at least one other feature.
We create 30 independent replications of this experiment by repeating the following process.
We first sample a gene $\rvx_1$ uniformly at random, adding it to the set $\mathcal{X}$.
For $\rvx_j$, $j > 1$, we sample $\rvx_k$ uniformly at random from $\mathcal{X}$ and compute the set of $50$ genes not in $\mathcal{X}$ with the highest correlation with $\rvx_k$.
From this set of $50$, we uniformly sample a gene $\rvx_j$ and add it to the feature set.
We repeat this process for $j = 2, \ldots, 100$, yielding $100$ genes in total.

We generate $\rvy \mid \rvx$ using a nonlinear response function adapted from a study on feature selection in neural network models of gene-drug interactions \citep{liang2018bayesian}.
The response consists of a nonlinear term, a second-order term, and a linear term.
For brevity, \cref{sec:liang-sim} contains the full simulation details.
We let \gls{ddlk} entropy regularization parameter $\lambda = 0.001$.

\begin{figure}[t]
\includegraphics[width=\textwidth,page=1,trim=0 0 0 0,clip]{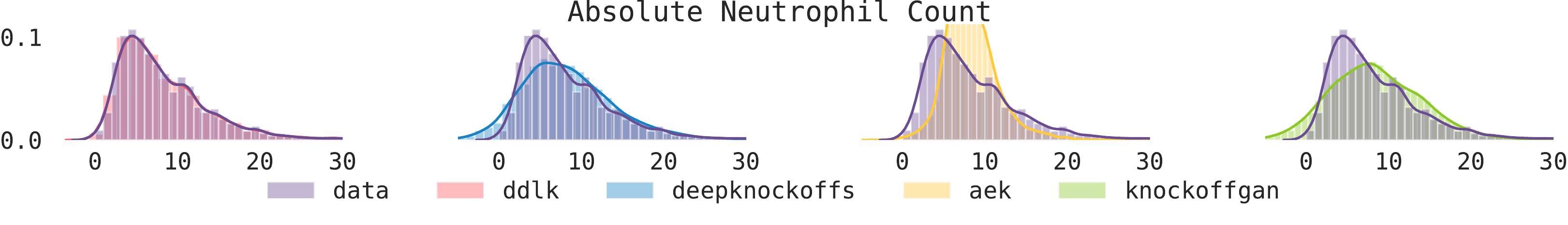}
\caption{
\textbf{\gls{ddlk} learns the marginals of COVID-19 data better than competing baselines.}
We plot the marginal distribution of a feature in a COVID-19 dataset, and the corresponding marginal of samples from each knockoff method.
}
\label{fig:covid-marginals-subset}
\vspace{-0.3cm}
\end{figure}

\paragraph{\emph{Results.}}

\Cref{fig:fdr-power-plots} (right) highlights the empirical \gls{fdp} and power of each knockoff generating method in the context of \texttt{gene}.
All methods control the \gls{fdr} below the nominal level, but the average \gls{fdp} of \gls{ddlk} at \gls{fdr} thresholds below 0.3 is closer to its expected value.
This range of thresholds is especially important as nominal levels of \gls{fdr} below 0.3 are most used in practice.
In this range, \gls{ddlk} achieves power on par with Deep Knockoffs at levels below 0.1, and higher power everywhere else.
Auto-Encoding Knockoffs and KnockoffGAN achieve noticeably lower power across all thresholds.
Deep Knockoffs perform well here likely due to a lack of strong third or higher moments of dependence between features.
We attribute the success of \gls{ddlk} and Deep Knockoffs to their ability to model highly correlated data.

\paragraph{COVID-19 adverse events.}

[\texttt{covid-19}]: The widespread impact of COVID-19 has led to the deployment of machine learning models to guide triage.
Data for COVID-19 is messy because of both the large volume of patients
and the changing practice for patient care.
Establishing trust in models for COVID-19 involves vetting the training data to ensure it does not contain artifacts that models can exploit.
Conditional independence tests help achieve this goal in two ways: (a) they highlight which features are most important to the response, and (b) they prune the feature set for a deployed model, reducing the risk of overfitting to processes in a particular hospital.
We apply each knockoff method to a large dataset from one of the epicenters of COVID-19 to understand the features most predictive of adverse events.

We use electronic health record data on COVID-positive patients from a large metropolitan health network.
Our covariates include demographics, vitals, and lab test results from every \gls{cbc} taken for each patient.
The response $\rvy \mid \rvx$ is a binary label indicating whether or not a patient had an adverse event (intubation, mortality, ICU transfer, hospice discharge, emergency department representation, O2 support in excess of nasal cannula at 6 L/min) within 96 hours of their \gls{cbc}.
There are 17K samples of 37 covariates in the training dataset, 5K in a validation set, and 6K in a held-out test set.
We let the \gls{ddlk} entropy regularization parameter $\lambda = 0.1$.
In this experiment, we use gradient boosted regression trees \citep{friedman2002stochastic, ke2017lightgbm} as our $\qresponse(\rvy \mid \rvx ; \gamma)$ model, and expected log-likelihood as a knockoff statistic.
We also standardize the data in the case of Deep Knockoffs since \glspl{mmd} that use the \gls{rbf} kernel with a single bandwidth parameter work better when features are on the same scale.

\begin{table}[t]
\centering
\tabcolsep=0.1cm
\begin{tabular}{@{}lcccc|c@{}}
\toprule
Feature             & \textbf{DDLK} & Deep Knockoffs  & AEK      & KnockoffGAN  & Validated \\ \midrule
Eosinophils count   & \cmark        & \xmark          & \xmark   & \xmark       & \cmark    \\
Eosinophils percent & \cmark        & \cmark          & \xmark   & \xmark       & \xmark    \\
Blood urea nitrogen & \cmark        & \xmark          & \xmark   & \xmark       & \cmark    \\
Ferritin            & \cmark        & \xmark          & \xmark   & \xmark       & \xmark    \\
O2 Saturation       & \cmark        & \xmark          & \xmark   & \xmark       & \cmark    \\
Heart rate          & \cmark        & \xmark          & \xmark   & \xmark       & \cmark    \\
Respiratory rate    & \cmark        & \cmark          & \xmark   & \xmark       & \cmark    \\
O2 Rate             & \cmark        & \cmark          & \cmark   & \cmark       & \cmark    \\
On room air         & \cmark        & \cmark          & \cmark   & \cmark       & \cmark    \\
High O2 support     & \cmark        & \cmark          & \cmark   & \cmark       & \cmark    \\
Age                 & \xmark        & \xmark          & \cmark   & \cmark       & \xmark    \\ \bottomrule
\end{tabular}
\vspace{0.1cm}
\caption{
\textbf{\Gls{ddlk} selects 10/37 features, 8 of which were found to be meaningful by doctors at a large metropolitan hospital.}
Here we show the union of \texttt{covid-19} features selected by each knockoff method at a nominal \gls{fdr} of 0.2.
Deep Knockoffs, Auto-Encoding Knockoffs, and KnockoffGAN exhibit lower power to select important features.
}
\label{tab:covid}
\vspace{-0.3cm}
\end{table}

\paragraph{\emph{Results.}}

As COVID-19 is a recently identified disease, there is no ground truth set of important features for this dataset.
We therefore use each knockoff method to help discover a set of features at a nominal \gls{fdr} threshold of $0.2$, and validate each feature by manual review with doctors at a large metropolitan hospital.
\Cref{tab:covid} shows a list of features returned by each knockoff method, and indicates whether or not a team of doctors thought the feature should have clinical relevance.

We note that \gls{ddlk} achieves highest power to select features, identifying 10 features, compared to 5 by DeepKnockoffs, and 4 each by Auto-Encoding Knockoffs and KnockoffGAN.
To understand why, we visualize the marginal distributions of each covariate in, and the respective marginal distribution of samples from each knockoff method, in \cref{fig:covid-marginals-subset}.

We notice two main differences between \gls{ddlk} and the baselines.
First, \gls{ddlk} is able to fit asymmetric distributions better than the baselines.
Second, despite the fact that the implementation of \gls{ddlk} using mixture density networks is misspecified for discrete variables, \gls{ddlk} is able to model them better than existing baselines.
This implementation uses continuous models for $\rvx$, but is still able approximate discrete distributions well.
The components of each mixture appear centered around a discrete value, and have very low variance as shown in \cref{fig:covid-marginals-ddlk}.
This yields a close approximation to the true discrete marginal.
We show the marginals of every feature for each knockoff method in \cref{sec:additional-figures}.

\vspace{-0.2cm}
\section{Conclusion}
\vspace{-0.2cm}
\Gls{ddlk} is a generative model for sampling knockoffs that directly minimizes a $\KL$ divergence implied by the knockoff swap property.
The optimization for \gls{ddlk} involves first maximizing the explicit likelihood of the covariates, then minimizing the $\KL$ divergence between the joint distribution of covariates and knockoffs and any swap between them.
To ensure \gls{ddlk} satisfies the swap property under any swap indices, we use the Gumbel-Softmax trick to learn swaps that maximize the $\KL$ divergence.
To generate knockoffs that satisfy the swap property while maintaining high power to select variables, \gls{ddlk} includes a regularization term that encourages high conditional entropy of the knockoffs given the covariates.
We find \gls{ddlk} to outperform various baselines on several synthetic and real benchmarks including a task involving a large dataset from one of the epicenters of COVID-19.

\section*{Broader Impact}
There are several benefits to using flexible methods for identifying conditional independence like \gls{ddlk}.
Practitioners that care about transparency have traditionally eschewed deep learning, but methods like \gls{ddlk} can present a solution.
By performing conditional independence tests with deep networks and by providing guarantees on the false discovery rate, scientists and practitioners can develop more trust in their models.
This can lead to greater adoption of flexible models in basic sciences, resulting in new discoveries and better outcomes for the beneficiaries of a deployed model.
Conditional independence can also help detect bias from data, by checking if an outcome like length of stay in a hospital was related to a sensitive variable like race or insurance type, even when conditioning on many other factors.

While we believe greater transparency can only be better for society, we note that interpretation methods for machine learning may not exactly provide transparency.
These methods visualize only a narrow part of a model's behavior, and may lead to gaps in understanding.
Relying solely on these domain-agnostic methods could yield unexpected behavior from the model.
As users of machine learning, we must be wary of too quickly dismissing the knowledge of domain experts.
No interpretation technique is suitable for all scenarios, and different notions of transparency may be desired in different domains.

\bibliographystyle{plainnat}
\bibliography{citations}

\newpage
\section*{Appendix}
\appendix
\section{Proofs}
\label{sec:proofs}

\subsection{Proof of \cref{thm:swap}}

Let $\mu$ be a probability measure defined on a measurable space.
Let $f_H$ be a swap function using indices $H \subseteq [d]$.
If $\vv$ is a sample from $\mu$, the probability law of $f_H(\vv)$ is $\mu \circ f_H$.

\begin{proof}
The swap operation $f_H$ on $[\rvx, \rvxt]$ swaps coordinates in the following manner:
for each $j \in H$, the $j$th and $(j + d)$th coordinates are swapped.
Let $(E, \mathcal{E})$ be a measurable space, where elements of $E$ are $2d$-dimensional vectors, and $\mathcal{E}$ is a $\sigma$-algebra on $E$.
Let $(F, \mathcal{F})$ also be a measurable space where each element of $F$ is an element of $E$ but with the $j$th coordinate swapped with the $(j+d)$th coordinate for each $j \in H$.
Similarly, let $\mathcal{F}$ be constructed by applying the same swap transformations to each element of $\mathcal{E}$.
$\mathcal{F}$ is a $\sigma$-algebra as swaps are one-to-one transformations, and $\mathcal{E}$ is a $\sigma$-algebra.

We first show that $f_H$ is a measurable function with respect to $\mathcal{E}$ and $\mathcal{F}$.
This is true by construction of the measurable space $(F, \mathcal{F})$.
For every element $B \in \mathcal{F}$, $f_H^{-1} (B) \in \mathcal{E}$.

We can now construct a mapping $\mu \circ f_H^{-1} (B)$ for all $B \in \mathcal{F}$.
This is the pushforward measure of $\mu$ under transformation $f_H$, and is well defined because $f_H$ is measurable.

Using the fact that a swap applied twice is the identity, we get $f_H = f_H^{-1}$.
With this, we see that the probability measure on $(F,\mathcal{F})$ is $\mu \circ f_H^{-1} = \mu \circ f_H$.
\end{proof}

\subsection{Alternative derivation of \cref{thm:swap} for continuous random variables}

In this section we derive \cref{thm:swap} for continuous random variables in an alternative manner.
Let $\rvx$ be a set of covariates, and $\rvxt$ be a set of knockoffs.
Let $\rvz = [\rvx, \rvxt]$, and $\rvw = [\rvx, \rvxt]_{\swap{H}}$ where $H$ is a set of coordinates in which we swap $\rvx$ and $\rvxt$.
Recall that a swap operation on $\rvz$ is an affine transformation $\rvw = \mathbf{A} \rvz$, where $\mathbf{A}$ is a permutation matrix.
Using this property, we get:
\begin{align*}
q_\rvw(\rvz) 
= \left | \text{det}\left(\frac{\partial \mathbf{A}^{-1} \rvw}{\partial \rvw}\right)\right |  \cdot q_{\rvz}(\mathbf{A}^{-1} \rvz) 
= q_{\rvz}(\mathbf{A}^{-1} \rvz)
= q_{\rvz}(\mathbf{A} \rvz)
= q_{\rvz}(\rvw).
\end{align*}
The first step is achieved by using a change of variables, noting that $\mathbf{A}$ is invertible, and $\rvz = \mathbf{A}^{-1} \rvw$.
The determinant of the Jacobian here is just the determinant of $\mathbf{A}^{-1}$.
$\mathbf{A}^{-1}$ is a permutation matrix whose parity is even, meaning its determinant is $1$, and that $\mathbf{A}^{-1} = \mathbf{A}$.
I.e. the density of the swapped variables evaluated at $\rvz$ is equal to the original density evaluated at $\rvw$.

\subsection{Proof of \cref{lemma:gumbel-correctness}}

Recall \cref{lemma:gumbel-correctness}:

\textbf{Worst case swaps}:
Let $q(H ; \beta)$ be the worst case swap distribution.
That is, the distribution over swap indices that maximizes
\begin{align*}
	\E_{H \sim q(H ; \beta)} \kl{\qjoint(\rvx ; \theta) \qknockoff(\rvxt \mid \rvx ; \phi)}{\qjoint(\rvu ; \theta) \qknockoff(\rvut \mid \rvu ; \phi)}
\end{align*}
with respect to $\beta$.
If this quantity is minimized with respect to $\phi$, knockoffs sampled from $\qknockoff$ will satisfy the swap property in \cref{eqn:swap} for any swap $H$ in the power set of $[d]$.

\begin{proof}
If
\begin{align}
	\E_{H \sim q(H ; \beta)} \kl{\qjoint(\rvx ; \theta) \qknockoff(\rvxt \mid \rvx ; \phi)}{\qjoint(\rvu ; \theta) \qknockoff(\rvut \mid \rvu ; \phi)} \label{eqn:worst-swap}
\end{align}
is minimized with respect to $\phi$ but maximized with respect to $\beta$, then for any other distribution $q(H ; \beta')$, \cref{eqn:worst-swap} will be lesser.
Minimizing \cref{eqn:worst-swap}, which is non-negative, with respect to $\phi$ implies that for any swap $H$ sampled from $q(H ; \beta)$ and for any knockoff $\rvxt$ sampled from $\qknockoff$,
\begin{align*}
	[\rvx, \rvxt] \overset{d}{=} [\rvx, \rvxt]_{\swap{H}}.
\end{align*}
As \cref{eqn:worst-swap} is also maximized with respect to $\beta$, swaps $H'$ drawn from all other distributions $q(H' ; \beta')$ will only result in lower values of \cref{eqn:worst-swap}.
Therefore, the joint distribution $[\rvx, \rvxt]$ will be invariant under any swap $H'$ in the power set of $[d]$.
\end{proof}

\subsection{Sufficient sets for swap condition expectation}
\label{sec:sufficient-swap}

Recall the swap property required of knockoffs highlighted in \cref{eqn:swap}:
\begin{align*}
	[\rvx, \rvxt] & \overset{d}{=} [\rvx, \rvxt]_{\text{swap}(H)}
\end{align*}
where $H \subseteq [d]$ is a set of coordinates under which we swap the covariates and knockoffs.
For valid knockoffs, this equality in distribution must hold for any such $H$.
One approach to check if knockoffs are valid is to verify this equality in distribution for all singleton sets $\{ j \} \subset [d]$ \citep{DeepKnockoffs,knockoffgan}.
To check if the swap property \cref{eqn:swap} holds under any $H = \{ j_1, \dots, j_k \}$, it suffices to check if \cref{eqn:swap} holds under each of $\{j_1 \}, \dots, \{ j_k \}$.

We can generalize this approach to check the validity of knockoffs under other collections of indices besides singleton sets using the following property.
Let $H_1, H_2 \subseteq [d]$ and 
\begin{align*}
	[\rvx, \rvxt] & \overset{d}{=} [\rvx, \rvxt]_{\swap{H_1}} \\
	[\rvx, \rvxt] & \overset{d}{=} [\rvx, \rvxt]_{\swap{H_2}}.
\end{align*}
Then,
\begin{align*}
	[\rvx, \rvxt]_{\text{swap}(H_1 \Delta H_2)}
	\overset{d}{=}
	\left [ [\rvx, \rvxt]_{\swap{H_1}} \right]_{\swap{H_2}}
	\overset{d}{=}
	\left [ \rvx, \rvxt \right]_{\swap{H_2}}
	\overset{d}{=}
	[\rvx, \rvxt]
\end{align*}
where $H_1 \Delta H_2$ is the symmetric difference of $H_1$ and $H_2$.
Swapping the indices in $H_1 \Delta H_2$ is equivalent to swapping the indices in $H_1$, then the indices in $H_2$.
If $\exists j \in H_1 \wedge j \in H_2$, swapping $j$ twice will negate the effect of the swap.

We can extend this property to $K$ sets and define sufficient conditions to check if the swap property holds.
Let $\{ A_k \}_{k=1}^K$ be a sequence of sets where each $A_k \subseteq [d]$.
Let
\begin{align*}
	A_1^* &= A_1 \\
	 \forall k \in [K], A_k^* &= A_k \Delta A_{k-1}^* \\
	A_K^* &= \{ j \}.
\end{align*}
Checking the swap property \cref{eqn:swap} under a sequence of swaps $\{ A_k \}_{k=1}^K$ is equivalent to checking \cref{eqn:swap} under the singleton set $\{ j  \}$.
Therefore, the swap property must also hold under the singleton set $\{ j \}$.

If collection of sets of swap indices $A$ contains a sub-sequence $\{ A_k \}_{k=1}^K$ such that their sequential symmetric difference is the singleton $\{ j \}$ for each $j \in [d]$, then a set of knockoffs that satisfies the swap property under each $A_k \in A$, will also satisfy the swap property under each singleton set, which is sufficient to generate valid knockoffs.

\section{Implicit reparameterization of mixture density networks}
\label{sec:implicit}

In our experiments, we decompose both $\qjoint$ and $\qknockoff$ via the chain rule:
\begin{align*}
	q(\rvx \mid \cdot) = q(\rvx_1 \mid \cdot) \prod_{j=2}^d q(\rvx_j \mid \cdot, \rvx_1, \cdots, \rvx_{j-1}).
\end{align*}
We model each conditional $q(\rvx_j \mid \cdot)$ using mixture density networks \cite{bishop:1994:mixture-density-networks} which take the form
\begin{align*}
	q(\rvx_j \mid \cdot) = \sum_{k=1}^K \pi_k(\cdot ; \psi_k) \mathcal{N}(\mu_k(\cdot ; \eta_k), \sigma_k^2(\cdot ; \omega_k)) 
\end{align*}
where functions $\{ \pi_k \}_{k=1}^K$, $\{ \mu_k \}_{k=1}^K$, and $\{ \sigma_k \}_{k=1}^K$ characterize a univariate gaussian mixture.
These parameters of these functions are $[\psi_1, \dots, \psi_K, \nu_1, \dots, \nu_K, \omega_1, \dots, \omega_K]$.

\paragraph{Fitting $\qjoint$.}

Let $\theta$, the parameters of $\qjoint$ contain parameters for every conditional $q(\rvx_j \mid \rvx_1, \dots, \rvx_{j-1})$.
The optimization of $\theta$ is straightforward:
\begin{align*}
	\theta = \argmax_\theta \mathcal{L}(\theta) = \argmax_\theta \frac{1}{N} \sum_{i=1}^N \log \qjoint(\vx^{(i)} ; \theta)
\end{align*}
only requires taking the derivative of $\mathcal{L}(\theta)$.

\paragraph{Fitting $\qknockoff$.}

Let $\phi$, the parameters of $\qknockoff$ contain parameters for every conditional $q(\rvxt_j \mid \rvx_1, \dots, \rvx_d, \rvxt_1, \dots, \rvxt_{j-1})$.
Recall the loss function $\mathcal{L}(\phi, \beta)$
\begin{align*}
	\mathcal{L}(\phi, \beta) = \E_{H \sim \qgumbel(\beta)}
	\E_{\vx \sim \D_N}
	\E_{\vxt \sim \qknockoff(\rvxt \mid \vx ; \phi)}
	\log \frac{\qjoint(\vx ; \theta) \qknockoff(\vxt \mid \vx ; \phi)^{1 + \lambda}}{\qjoint(\vu ; \theta) \qknockoff(\vut \mid \vu ; \phi)}.
\end{align*}
The optimization of $\phi$ requires $\nabla_\phi \mathcal{L}(\phi, \beta)$, which involves the
derivative of an expectation with respect to to $\qknockoff(\rvxt \mid \rvx ; \phi)$.
We use implicit reparameterization \citep{figurnov2018implicit}.
The advantage of implicit reparameterization over explicit reparameterization \citep{kingma2015variational} is that an inverse standardization function $S_\phi^{-1}$ -- which transforms random noise into samples from a distribution parameterized by $\phi$ -- is not needed.
Using implicit reparameterization, gradients of some objective $\E_{q(\rvz ; \phi)} [f(\rvz)]$ can be rewritten as
\begin{align*}
	\E_{q(\rvz ; \phi)} [\nabla_\phi f(\rvz)] &= \E_{q(\rvz ; \phi)} [\nabla_\rvz f(\rvz) \nabla_\phi \rvz] \\
	&= \E_{q(\rvz ; \phi)} [ -\nabla_\rvz f(\rvz) 
	(\nabla_\rvz S_\phi(\rvz))^{-1} \nabla_\phi S_\phi(\rvz)
	].
\end{align*}

We use this useful property to reparameterize gaussian mixture models.
Let $q(\rvz ; \phi)$ be a gaussian mixture model:
\begin{align*}
	q(\rvz ; \phi) = \sum_{k=1}^K \pi_k \mathcal{N}(\rvz ; \mu_k, \sigma_k^2)
\end{align*}
where $\phi = [\pi_1, \dots, \pi_K, \mu_1, \dots, \mu_K, \sigma_1, \dots, \sigma_K]$.
Let the standardization function $S_\phi$ be the CDF of $q(\rvz ; \phi)$:
\begin{align*}
	S_\phi(\rvz) = \sum_{k=1}^K \pi_k \Phi \left( \frac{\rvz - \mu_k}{\sigma_k} \right)
\end{align*}
where $\Phi$ is the standard normal gaussian CDF.
We use this to compute the gradient of $\rvz$ with respect to each parameter:
\begin{align*}
	\nabla_{\pi_k} \rvz &= -\frac{\Phi(\frac{\rvz - \mu_k}{\sigma_k})}{q(\rvz ; \phi)} \\
	\nabla_{\mu_k} \rvz &= \frac{\pi_k \cdot \mathcal{N}(\rvz ; \mu_k, \sigma_k^2)}{q(\rvz ; \phi)} \\
	\nabla_{\sigma_k} \rvz &= \frac{\pi_k \cdot
	\left( \frac{\rvz - \mu_k}{\sigma_k} \right) \cdot
	\mathcal{N}(\rvz ; \mu_k, \sigma_k^2)}{q(\rvz ; \phi)}.
\end{align*}

Putting it all together, we use the implicit reparameterization trick to implement each conditional distribution in $\qjoint$ and $\qknockoff$.

\section{Implementation details and hyperparameter settings for \gls{ddlk} experiments}
\label{sec:hyperparameters}

We decompose $\qjoint$ and $\qknockoff$ into the product of univariate conditional distributions using the product rule.
We use mixture density networks \citep{bishop:1994:mixture-density-networks} to parameterize each conditional distribution.
Each mixture density network is a 3-layer neural network with 50 parameters in each layer and a residual skip connection from the input to the last layer.
Each network outputs the parameters for a univariate gaussian mixture with 5 components.
We initialize the network such that the modes are evenly spaced within the support of training data.

Using $\qgumbel$, we sample binary swap matrices of the same dimension as the data.
As we require discrete samples from the Gumbel-Softmax distribution, we implement a straight-through estimator \citep{gumbel-jang}.
The straight-through estimator facilitates sampling discrete indices, but uses a continuous approximation during backpropagation.

The $\qjoint$ model is optimized using Adam \citep{kingma2014adam}, with a learning rate of $5 \times 10^{-4}$ for a maximum of 50 epochs.
The $\qknockoff$ model is optimized using Adam, with a learning rate of $1 \times 10^{-3}$ for $\phi$ and $1 \times 10^{-2}$ for $\beta$ for a maximum of 250 epochs.
We also implement early stopping using validation loss using the PyTorch Lightning framework \citep{falcon2019pytorch}.

Our code can be found online by installing:
\begin{align*}
\texttt{pip install -i https://test.pypi.org/simple/ ddlk==0.2}
\end{align*}

\paragraph{Compute resources.}

We run each experiment on a single CPU core using 4GB of memory.
Fitting $\qjoint$ for a 100-dimensional dataset with 2000 samples requires fitting 100 conditional models, and takes roughly 10 minutes.
Fitting $\qknockoff$ for the same data takes roughly 30 minutes.

Fitting \gls{ddlk} to our \texttt{covid-19} dataset takes roughly 15 minutes in total.

\section{Baseline model settings}
\label{sec:baseline-model-settings}

For Deep Knockoffs and KnockoffGAN, we use code from each respective repository:
\begin{align*}
	&\texttt{https://github.com/msesia/deepknockoffs} \\
	&\texttt{https://bitbucket.org/mvdschaar/mlforhealthlabpub/}
\end{align*}
and use the recommended hyperparameter settings. 

At the time of writing this paper, there was no publicly available implementation for Auto-Encoding Knockoffs.
We implemented Auto-Encoding Knockoffs with a \gls{vae} with a gaussian posterior $q(\rvz \mid \rvx) \approx \mathcal{N}(\rvz; \mu_\rvz(\rvx), \sigma_\rvz(\rvx))$ and likelihood $p(\rvx \mid \rvz) \approx \mathcal{N}(\rvx; \mu_\rvx(\rvz), \sigma_\rvx(\rvz))$.
Each of $\mu_\rvz, \sigma_\rvz, \mu_\rvx. \sigma_\rvx$ is a 2-layer neural network with 400 units in the first hidden layer, 500 units in the second, and ReLU activations.
The outputs of networks $\sigma_\rvz, \sigma_\rvx$ are exponentiated to ensure variances are non-negative.
The outputs of network $\mu_\rvz$ and $\sigma_\rvz$ are of dimension $d_\rvz$, and the outputs of $\mu_\rvx$ and $\sigma_\rvx$ are of dimension $d$, the covariate dimension.
For each dataset, we choose the dimension $d_\rvz$ of latent variable $\rvz$ that maximizes the estimate of the ELBO on a validation dataset.
In our experiments, we search for $d_\rvz$ over the set $\{d_\rvz  : 10 \leq d_\rvz \leq 200, d_\rvz \mod 10 = 0 \}$.
For each dataset, we use the following $d_\rvz$:
\begin{enumerate}
	\item \texttt{gaussian}: 20
	\item \texttt{mixture}: 140
	\item \texttt{gene}: 30
	\item \texttt{covid-19}: 60.
\end{enumerate}
The neural networks are trained using Adam \citep{kingma2014adam}, with a learning rate of $1 \times 10^{-4}$ for a maximum of 150 epochs.
To avoid very large gradients, we standardize the data using the mean and standard deviation of the training set.
To generate knockoffs $\rvxt$, we use the same approach prescribed by \citet{liu2018auto}.
We first sample the latent variable $\rvz$ conditioned on the covariates using the posterior distribution:
\begin{align*}
	\rvz \sim \mathcal{N}\left(\rvz; \mu_\rvz(\rvx), \sigma_\rvz(\rvx) \right).
\end{align*}
This sample of $\rvz$ is then used to sample a knockoff $\rvxt$ using the likelihood distribution:
\begin{align*}
	\rvxt \sim \mathcal{N}\left(\rvx; \mu_\rvx(\rvz), \sigma_\rvx(\rvz) \right).
\end{align*}
Since these $\rvxt$ are standardized, we re-scale them by the the training mean and standard deviation.

\section{Robust model-based statistics for \gls{fdr}-control}
\label{sec:mixture-statistic}

The goal of any knockoff method is to help compute test statistics for a conditional independence test.
We employ a variant of \glspl{hrt} \citep{tansey2018holdout} to compute test statistics $w_j$ for each feature $\rvx_j$.
We split dataset $\D_N := \{ (\vx^{(i)}, \vy^{(i)}) \}_{i=1}^N$ into train and test sets $\D_N^{\text{(tr)}}$, and $\D_N^{\text{(te)}}$ respectively, then sample knockoff datasets $\Dnt^{\text{(tr)}}$ and $\Dnt^{\text{(te)}}$ conditioned on each.
Next, a model $\qresponse$ is fit with $\D_N^{\text{(tr)}}$.

To compute knockoff statistics with $\qresponse$, we use a measure of performance $\mathcal{W}(\qresponse, \D_N^{\text{(te)}})$ on the test set.
For real-valued $\rvy$, $\mathcal{W}$ is the mean squared-error, and for categorical $\rvy$, $\mathcal{W}$ is expected log-probability of $\rvy \mid \rvx$.
A knockoff statistic $w_j := \mathcal{W}(\qresponse, \D_N^{\text{(te)}}) - \mathcal{W}(\qresponse, \Djnt^{\text{(te)}})$ is recorded for each feature $\rvx_j$, where $\Djnt^{\text{(te)}}$ is $\D_N^{\text{(te)}}$ but with the $j$th feature swapped with $\Dnt^{\text{(te)}}$.

In practice, we use use flexible models like neural networks or boosted trees for $\qresponse$.
While the model-based statistic above will satisfy the properties detailed in \cref{sec:kfilters} and control \gls{fdr}, its ability to do so is hindered by imperfect knockoffs.
In such cases, we observe that knockoff statistics for null features are centered around some $\zeta > 0$, violating a condition required for empirical \gls{fdr} control.
This happens because if the covariates and knockoffs are not equal in distribution, 
models trained on the covariates will fit the covariates better than the knockoffs 
and inflate the value of test statistic $w_j$.
This can lead to an increase in the false discovery rate as conditionally independent features may be selected if their statistic is larger than the selection threshold.
To combat this, we propose a mixture statistic that trades off power for \gls{fdr}-control.

The mixture statistic involves fitting a $\qresponse$ model for each feature $\rvx_j$ using an equal mixture of data in $\D_N^{\text{(tr)}}$ and $\Djnt^{\text{(tr)}}$, then computing $\mathcal{W}$ as above.
Such a $\qresponse$ achieves lower performance on $\D_N^{\text{(te)}}$, but higher performance on $\Djnt^{\text{(te)}}$, yielding values of $w_j$ with modes closer to $0$, enabling finite sample \gls{fdr} control.
However, this \gls{fdr}-control comes at the cost of power as the method's ability to identify conditionally dependent features is reduced.

\section{Nonlinear response for \texttt{gene} experiments}
\label{sec:liang-sim}

We simulate the response $\rvy \mid \rvx$ for the \texttt{gene} experiment using a nonlinear response function designed for genomics settings \citep{liang2018bayesian}.
The response consists of two first-order terms, a second-order term, and an additional nonlinearity in the form of a $\tanh$:
\begin{align*}
		k &\in [m / 4] \\
		\varphi^{(1)}_k, \varphi^{(2)}_k &\sim \mathcal{N}(1, 1) \\
		\varphi^{(3)}_k, \varphi^{(4)}_k, \varphi^{(5)}_k, \varphi^{(6)}_k &\sim \mathcal{N}(2, 1)
\end{align*}
\begin{align*}
	\rvy \mid \rvx &= \epsilon + \sum_{k=1}^{m / 4} 
	\varphi^{(1)}_k \rvx_{4k - 3} + 
	\varphi^{(3)}_k \rvx_{4k - 2} + 
	\varphi^{(4)}_k \rvx_{4k - 3} \rvx_{4k - 2} + 
	\varphi^{(5)}_k \tanh (\varphi^{(2)}_k \rvx_{4k - 1} + \varphi^{(6)}_k \rvx_{4k})
\end{align*}
where $m$ is the number of important features.
In our experiments, we set $m = 20$.
This means that the first $20$ features are important, while the remaining $80$ are unimportant.

\section{Generating knockoffs for COVID-19 data}
\label{sec:additional-figures}

\begin{figure}[t]
    \makebox[\textwidth][l]{
    \includegraphics[width=0.92\textwidth,page=4]{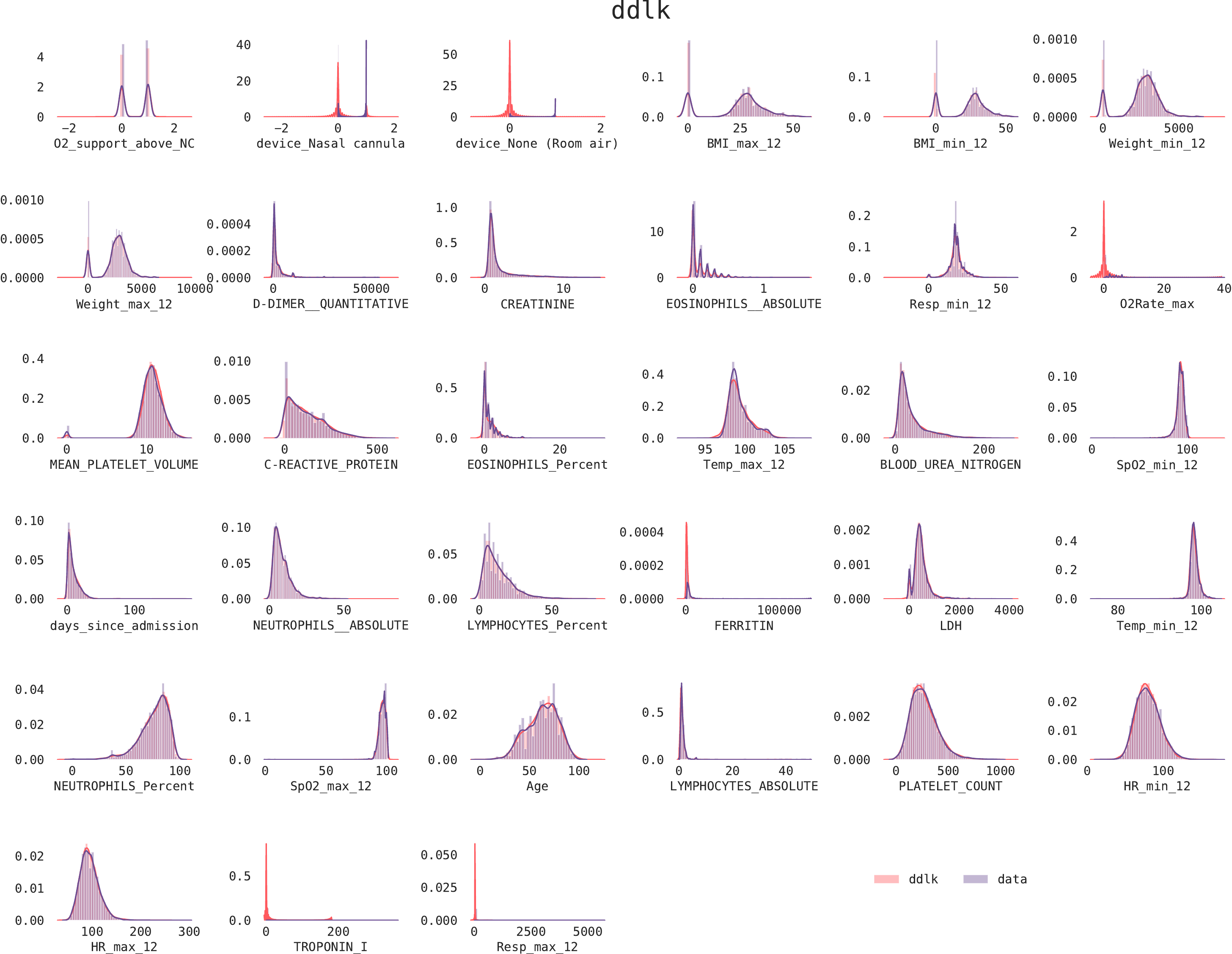}
    }
    
    \caption{\small{\textbf{The marginal distributions of knockoff samples from \gls{ddlk} look very similar to those from the data.} Despite this implementation of \gls{ddlk} using mixture density networks, the modes of each marginal line up with discrete values in the data.} \normalsize
    }
    
    \label{fig:covid-marginals-ddlk}
\end{figure}

\begin{figure}[t]
    \makebox[\textwidth][l]{
    \includegraphics[width=0.92\textwidth,page=3]{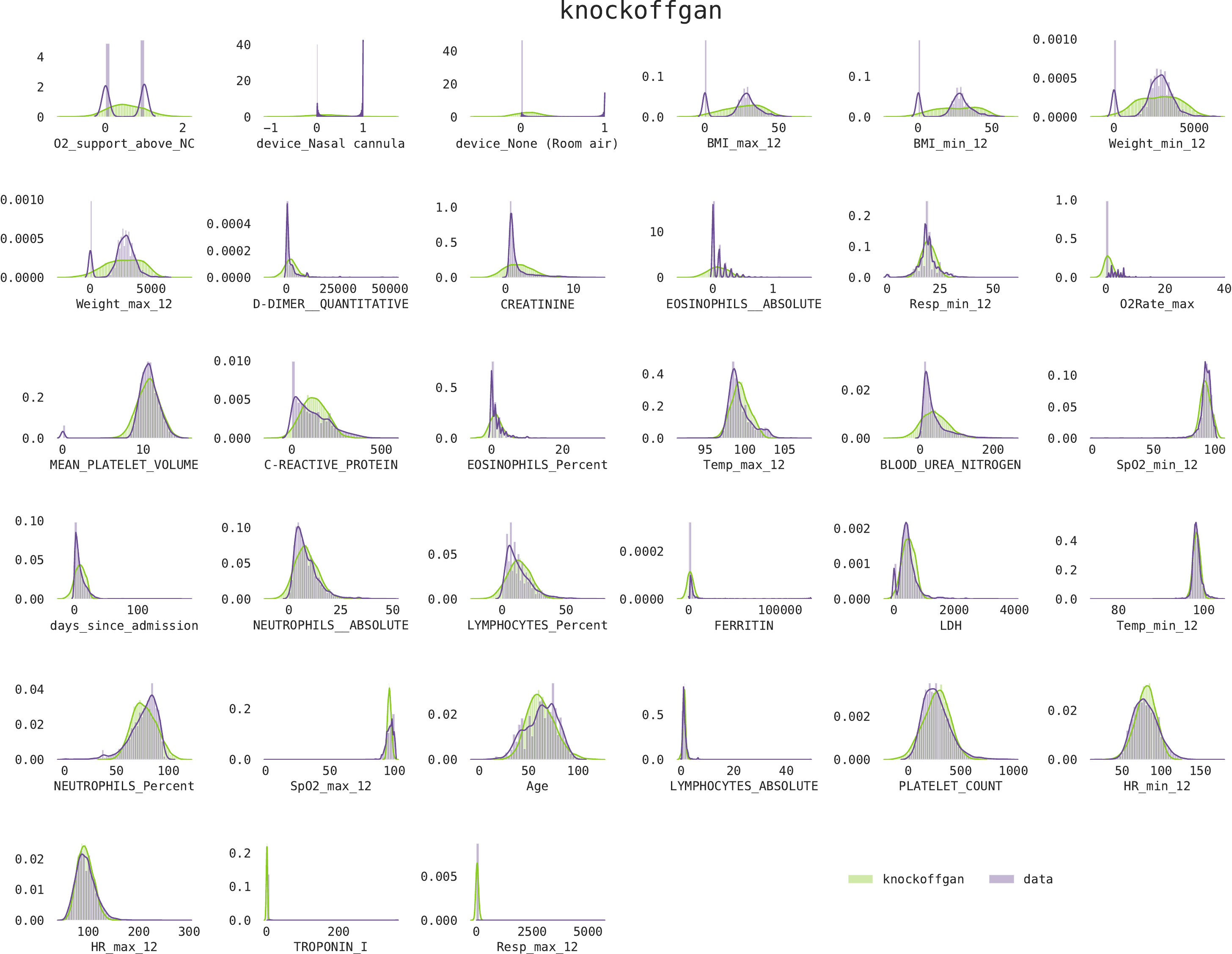}
    }
    \caption{\small {The marginals distributions of samples from KnockoffGAN match the data only when the feature $\rvx_j$ is univariate, and has roughly equal mass on either side of the mode.} \normalsize}
    \label{fig:covid-marginals-knockoffgan}
\end{figure}

\begin{figure}[t]
    \makebox[\textwidth][l]{
    \includegraphics[width=0.92\textwidth,page=2]{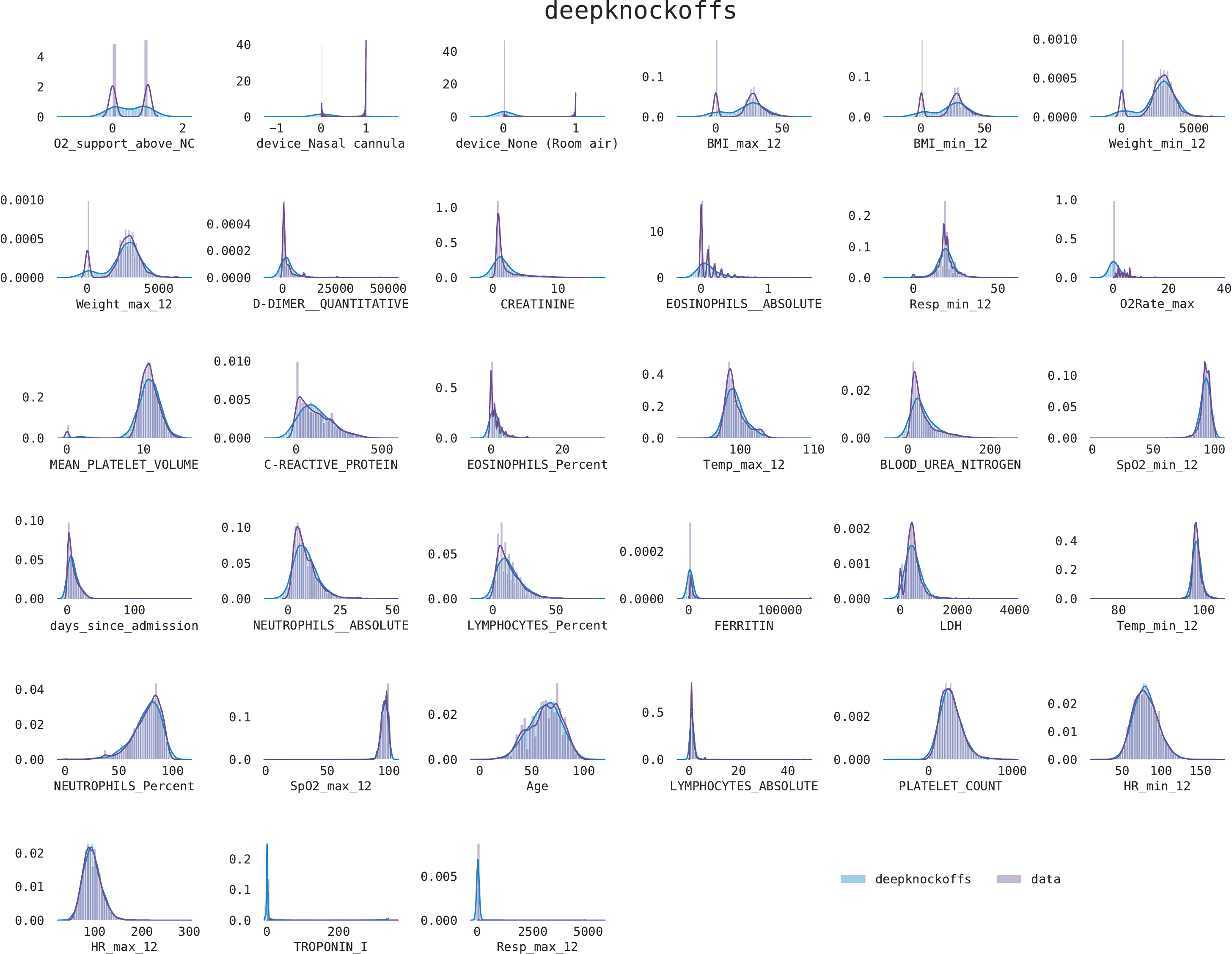}
    }
    \caption{The marginals distributions of samples from Deep Knockoffs match the data only when the feature $\rvx_j$ is univariate and has fat tails.}
    \label{fig:covid-marginals-deepknockoffs}
\end{figure}

\begin{figure}[t]
    \makebox[\textwidth][l]{
    \includegraphics[width=0.92\textwidth,page=1]{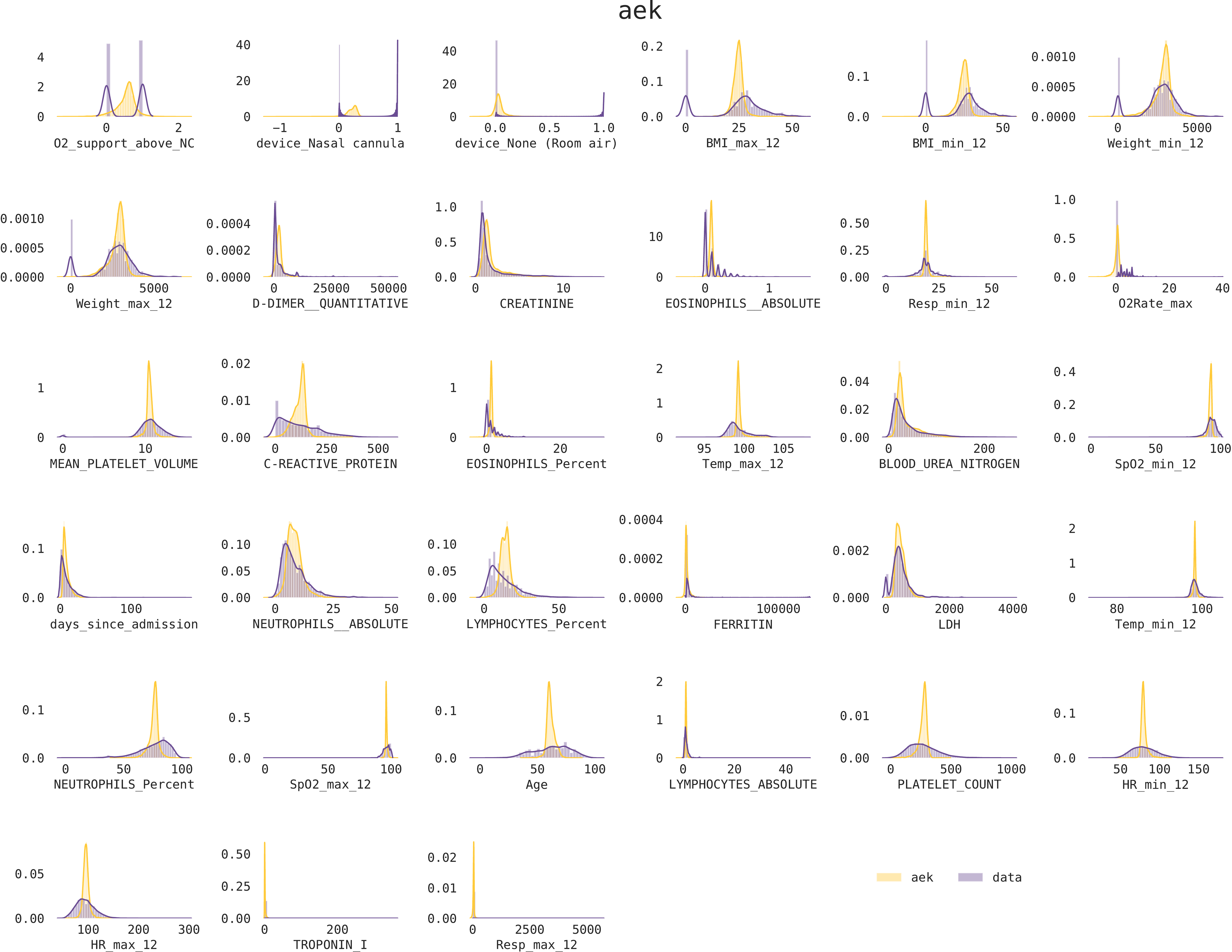}
    }
    \caption{Auto-Encoding Knockoffs tend to learn underdispersed distributions for the covariates. Further, all of the marginal distributions learned are univariate and exhibit variance much smaller than that of the data.}
    \label{fig:covid-marginals-aek}
\end{figure}

In this section, we visualize the marginals of each feature in our \texttt{covid-19} dataset, and the marginals of knockoffs sampled from each method.
\Cref{fig:covid-marginals-ddlk,fig:covid-marginals-knockoffgan,fig:covid-marginals-deepknockoffs,fig:covid-marginals-aek} plot the marginals of samples from \gls{ddlk}, KnockoffGAN, Deep Knockoffs, and Auto-Encoding Knockoffs respectively.
These provide insight into why \gls{ddlk} is able to select more features at the same nominal \gls{fdr} threshold of 0.2.
We first notice that knockoff samples from \gls{ddlk} match the marginals of the data very well.
\Gls{ddlk} is the only method that models asymmetric distributions well.

Despite our experiment using mixture density networks to implement \gls{ddlk}, discrete data is also modeled well.
For example the values of \texttt{O2\_support\_above\_NC} -- a binary feature indicating whether a patient required  oxygen support greater than nasal cannula -- are also the modes of a mixture density learned by \gls{ddlk}.
Samples from Auto-Encoding Knockoffs, KnockoffGAN, and Deep Knockoffs tend to place mass spread across these values.

\end{document}